\documentclass[12pt]{article}

\usepackage[
  margin=1.5cm,
  includefoot,
  footskip=30pt,
]{geometry}
\usepackage[table]{xcolor}
\usepackage{graphicx}
\usepackage{tikz}
\usetikzlibrary{matrix, arrows}
\usepackage{amsmath,amssymb}
\usepackage{amsthm}
\usepackage{mathtools}
\usepackage{xspace}
\usepackage[noend]{algorithmic}
\usepackage[ruled,vlined]{algorithm2e}
\usepackage{url}
\usepackage{makeidx}
\usepackage{enumerate}
\usepackage{epstopdf}
\usepackage{booktabs}
\usepackage{color}
\usepackage[utf8]{inputenc}
\usepackage{thm-restate}
\usepackage{scalerel,stackengine}
\usepackage[shortlabels]{enumitem}
\usepackage{xr}
\usepackage{fancyvrb}
\usepackage{bold-extra}
\usepackage{subcaption}
\usepackage[width=474.18663pt]{caption}
\usepackage[most]{tcolorbox}
\usepackage{fvextra}
\usepackage{float}
\usepackage{alltt}
\usepackage{soul}
\usepackage{fancyvrb}
\usepackage{multirow}
\usepackage{array}
\usepackage{threeparttable}
\usepackage{authblk}
\usepackage{multicol}
\usepackage{xcolor}
\usepackage{colortbl}
\usepackage{siunitx}
\usepackage{fancyvrb}
\usepackage[final]{hyperref}
\usepackage[bottom]{footmisc}
\usepackage{makecell} 
\usepackage{fancyhdr} %

\newcommand{\eat}[1]{\ignorespaces}

\usepackage{listings}
\usepackage{tikz}
\usetikzlibrary{shapes,calc,positioning}

\global

\tcbset{
  aibox/.style={
    width=474.18663pt,
    top=10pt,
    colback=white,
    colframe=black,
    colbacktitle=black,
    enhanced,
    center,
    attach boxed title to top left={yshift=-0.1in,xshift=0.15in},
    boxed title style={boxrule=0pt,colframe=white,},
  }
}
\newtcolorbox{AIbox}[2][]{aibox,title=#2,#1}

\definecolor{Gray}{gray}{0.95}
\definecolor{aigold}{RGB}{244,210, 1} 
\definecolor{aigreen}{RGB}{210,244,211} 

\sethlcolor{aigreen}

\definecolor{aired}{RGB}{255,180,181}

\newtcbox{\mybox}[1][green]{on line,
arc=0pt,outer arc=0pt,colback=#1!10!white,colframe=#1!50!black,
boxsep=0pt,left=0pt,right=0pt,top=0pt,bottom=0pt,
boxrule=0pt,bottomrule=0pt,toprule=0pt}

\title{From Medprompt to o1: Exploration of Run-Time Strategies for Medical Challenge Problems and Beyond}

\author[ ]{
    \centering
    Harsha Nori{\textsuperscript{*}\textsuperscript{\textdaggerdbl}}, Naoto Usuyama\textsuperscript{*} \\
    Nicholas King, Scott Mayer McKinney\textsuperscript{\S}, Xavier Fernandes, Sheng Zhang, \\
    Eric Horvitz\textsuperscript{\textdaggerdbl}
}
\date{}

\affil[$~$]{Microsoft}
\affil[$\S$]{OpenAI}

\begin{document}

\maketitle

\renewcommand{\thefootnote}{*}
\footnotetext{These authors contributed equally.}

\renewcommand{\thefootnote}{\textdaggerdbl}
\footnotetext{Correspondence: hanori@microsoft.com, horvitz@microsoft.com}

\renewcommand{\thefootnote}{\arabic{footnote}}

\begin{abstract}

Run-time steering strategies like Medprompt are valuable for guiding large language models (LLMs) to top performance on challenging tasks. Medprompt demonstrates that a general LLM can be focused to deliver state-of-the-art performance on specialized domains like medicine by using a prompt to elicit a run-time strategy involving chain of thought reasoning and ensembling. OpenAI's o1-preview model represents a new paradigm, where a model is designed to do run-time reasoning before generating final responses. We seek to understand the behavior of o1-preview on a diverse set of medical challenge problem benchmarks. Following on the Medprompt study with GPT-4, we systematically evaluate the o1-preview model across various medical benchmarks. Notably, even without prompting techniques, o1-preview largely outperforms the GPT-4 series with Medprompt. We further systematically study the efficacy of classic prompt engineering strategies, as represented by Medprompt, within the new paradigm of reasoning models. We found that few-shot prompting hinders o1's performance, suggesting that in-context learning may no longer be an effective steering approach for reasoning-native models. While ensembling remains viable, it is resource-intensive and requires careful cost-performance optimization. Our cost and accuracy analysis across run-time strategies reveals a Pareto frontier, with GPT-4o representing a more affordable option and o1-preview achieving state-of-the-art performance at higher cost. Although o1-preview offers top performance, GPT-4o with steering strategies like Medprompt retains value in specific contexts. Moreover, we note that the o1-preview model has reached near-saturation on many existing medical benchmarks, underscoring the need for new, challenging benchmarks. We close with reflections on general directions for inference-time computation with LLMs.

\end{abstract}

\newpage

\section{Introduction}

Prompt engineering as a research area and craft has evolved in step with the fast-paced rise of applications of large language models (LLMs). Prompts shape and focus the capabilities of LLMs trained to follow instructions. In our previous work, we introduced Medprompt, highlighting the effectiveness of inference-time capabilities through use of a structured, multi-step prompt pipeline. Medprompt, developed through exploratory work on prompting strategies to enhance model performance on medical challenge benchmarks, significantly boosts performance by leveraging dynamic chain-of-thought reasoning, curated few-shot examples, and choice-shuffle ensembling \cite{nori2023can}. We found that these methods focus and amplify the reasoning abilities of LLMs, with particularly valuable application in high-stakes domains like medical diagnostics and decision-making.

\begin{figure}[H]
    \centering
    \includegraphics[width=\textwidth]{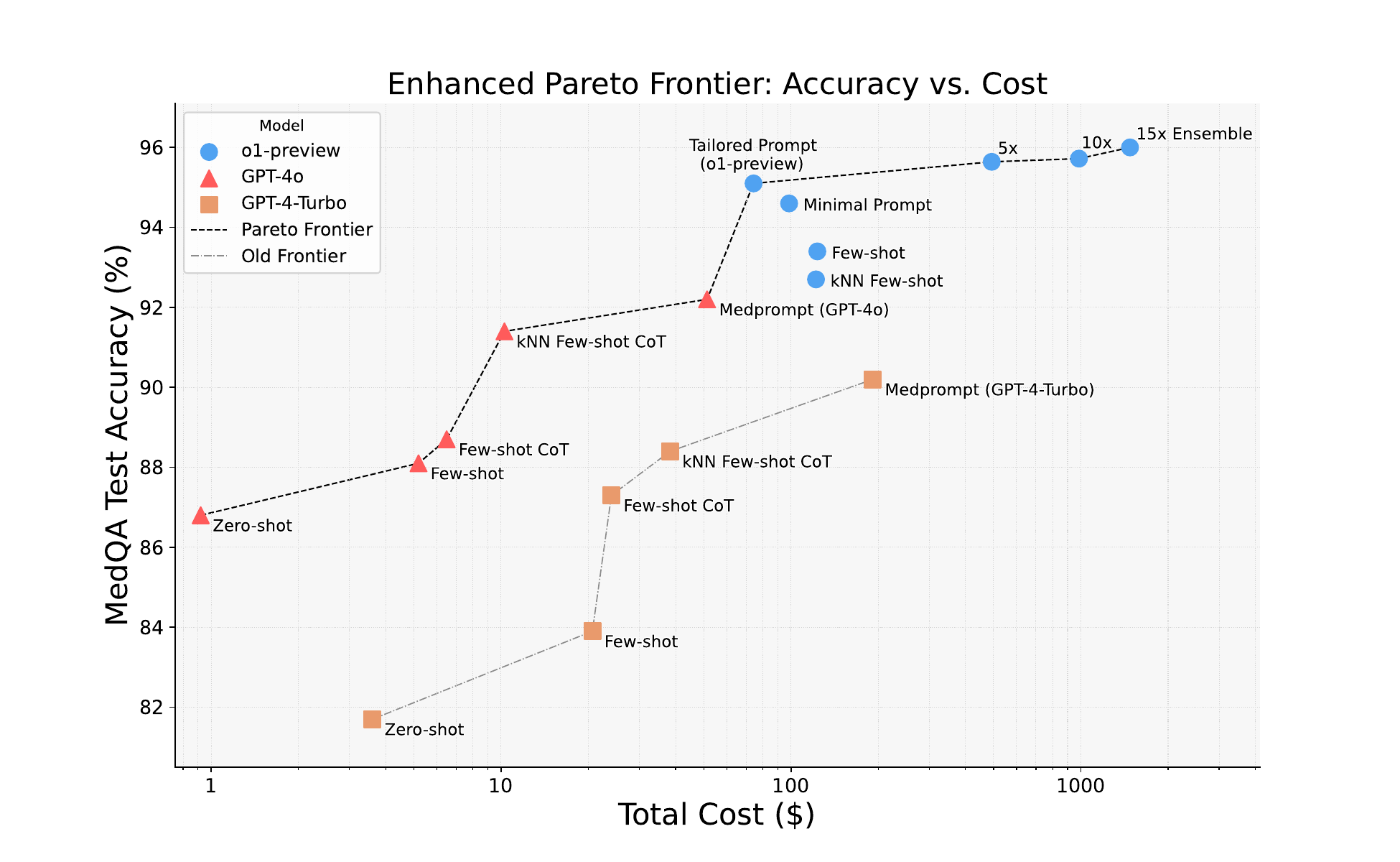}
    \caption{Pareto frontier showing accuracy versus total API cost (log scale) on the MedQA benchmark (1273 questions total). We compare o1-preview (Sep 2024), GPT-4o (Aug 2024), and GPT-4 Turbo (Nov 2023) with various run-time steering strategies.}
    \label{fig:cost}
\end{figure}

Medprompt shows that inference-time strategies can bridge the gap between general-purpose LLMs and domain-specific models that rely on a fixed set of expert-curated prompts and fine-tuning on specialized datasets. In particular, Medprompt demonstrates how error rates on complex medical benchmarks such as MedQA can be reduced by nearly 50\%, without adapting model weights to the medical domain.

Recent advancements in model training methodologies, exemplified by OpenAI's o1-preview model, represent a  novel approach to harnessing the inherent capabilities of LLMs. In distinction to previous models, o1-preview incorporates chain-of-thought (CoT) reasoning as part of its training process, yielding ``reasoning-native'' models that inherently perform sophisticated step-by-step problem-solving during inference. The integration of such intrinsic inference capabilities potentially reduces the need for extensive prompt engineering techniques aimed at extracting maximum performance.

\begin{figure}[H]
\vspace{10pt}
    \centering
    \begin{subfigure}[t]{0.49\textwidth}
        \centering
        \includegraphics[width=1\textwidth]{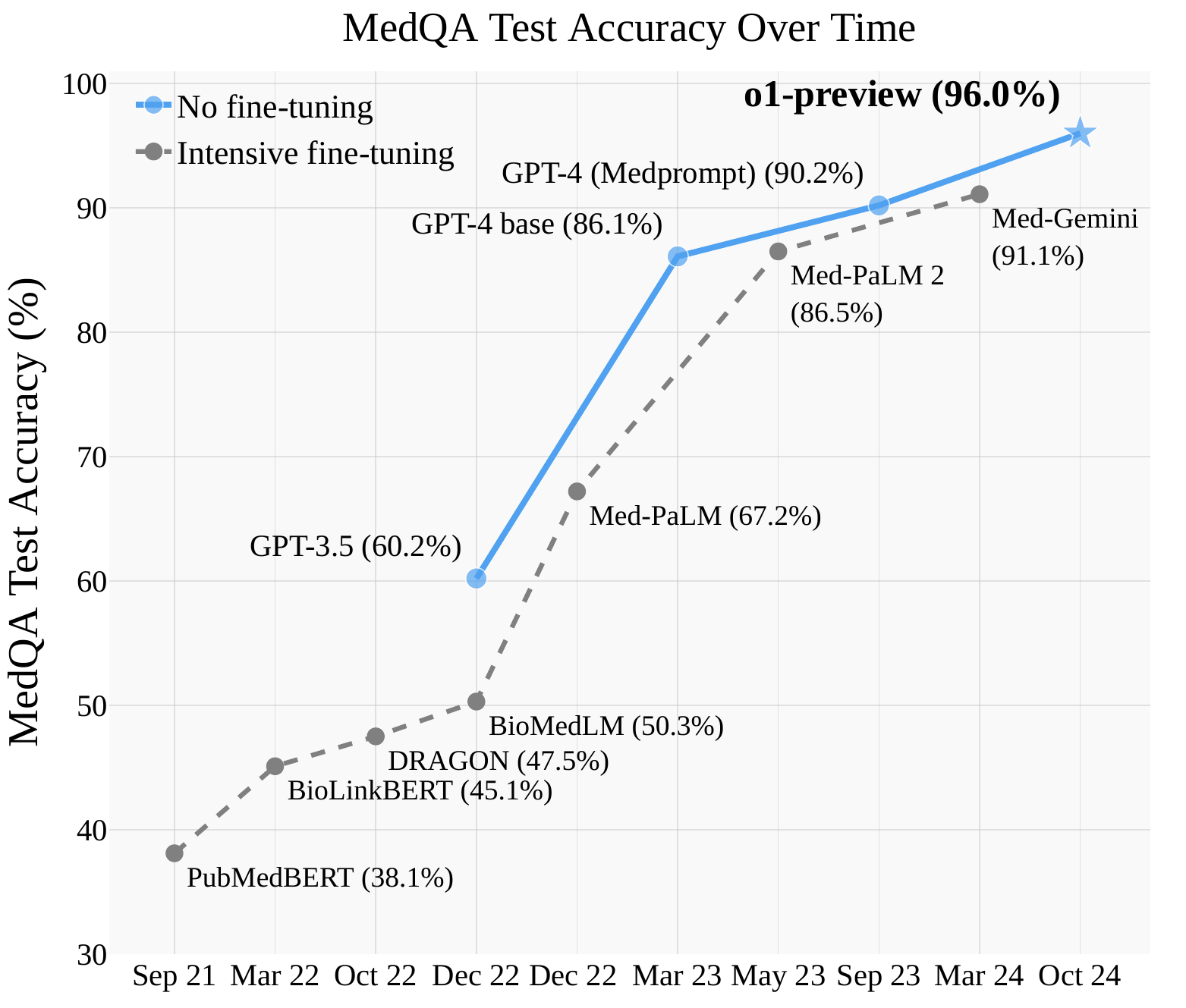}
        \caption{}
        \label{subfig:medqa-comp}
    \end{subfigure}
    \hfill
    \begin{subfigure}[t]{0.49\textwidth}
        \centering
        \includegraphics[width=1\textwidth]{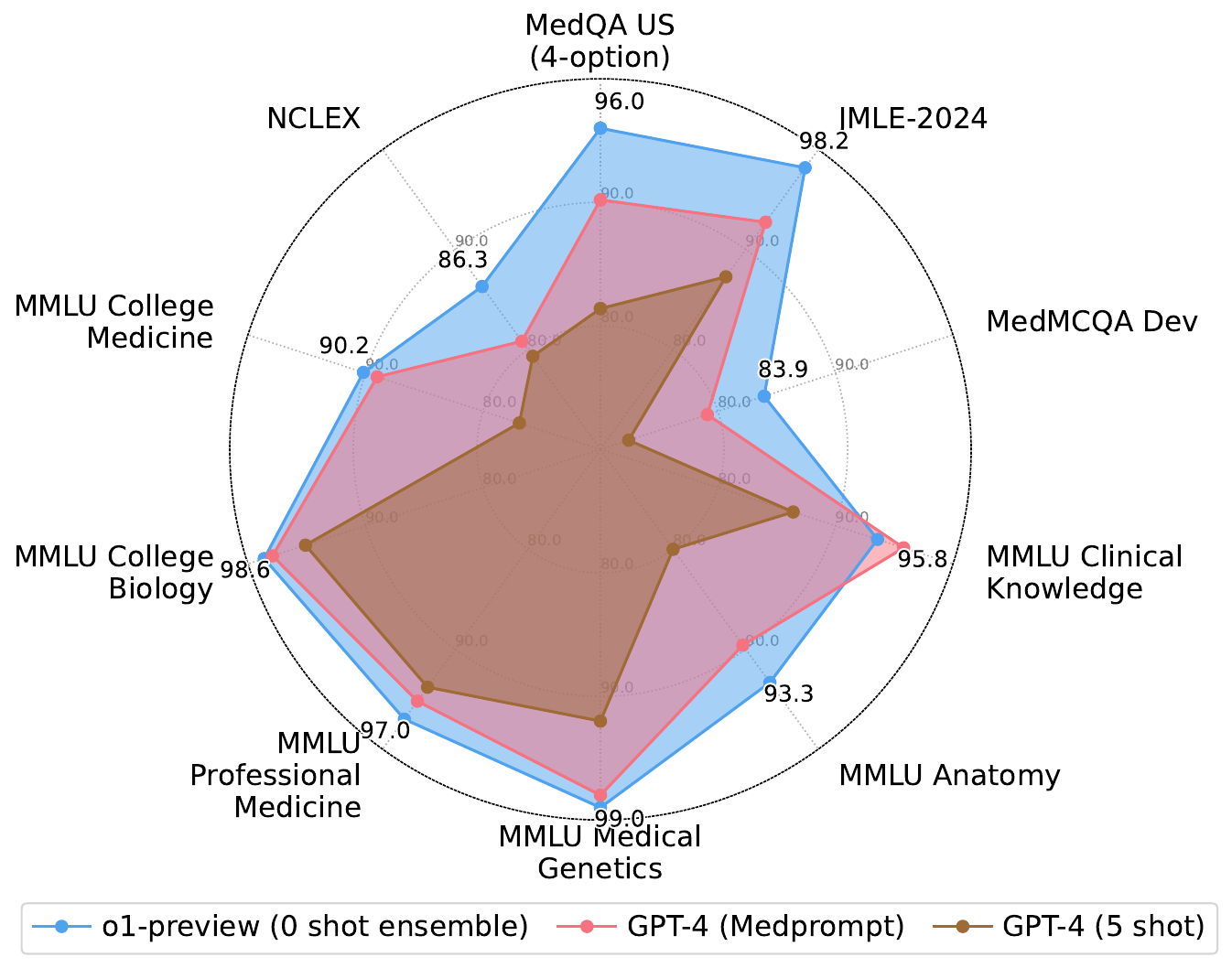}
        \caption{}
        \label{subfig:main-radar}
    \end{subfigure}
    \caption{(a) Comparative analyses of performance of multiple models on MedQA. (b) Comparisons on a wide range of medical challenge benchmarks.}
    \label{fig:comparison}
\end{figure}

We re-examine the role and value of sophisticated prompt-engineering strategies, as represented by Medprompt, given the advent of a new paradigm of models that perform run-time CoT reasoning, as represented by the o1 series. Our findings (Figure~\ref{fig:cost} and \ref{fig:comparison}) indicate that the o1-preview model outperforms GPT-4 augmented with Medprompt on the benchmarks studied and suggest a diminishing necessity for elaborate prompt-engineering techniques that were highly advantageous for earlier generations of LLMs. These results underscore an evolving landscape in which advances in model training have begun to internalize the principles behind prompt engineering aimed at run-time optimization.

\section{Background}
\subsection{Leveraging LLMs for medical challenge problems}

We focus on benchmarks in medical challenge problems as a representative specialist area for studying the use of generalist models in realms that rely on highly trained specialists. Generalist LLMs have demonstrated remarkable capabilities for challenging problems in the health and life sciences, often surpassing models constructed with significant domain-specific training and fine-tuning~\cite{nori2023capabilities,singhal2022large}. 

In the early era of foundation models, domain-specific pretraining was crucial due to their limited size and computational power. Models such as PubMedBERT~\cite{pubmedbert}, BioGPT~\cite{biogpt}, and BioMedLM~\cite{bolton2024biomedlm} were pretrained on specialized datasets like the PubMed corpus and the UMLS knowledge graph using self-supervised learning. Despite their relatively small sizes, these models delivered strong performance on biomedical NLP tasks. However, recent advancements have shown that newer, larger general-domain foundation models can achieve superior performance in medical challenges, even without the need for domain-specific pretraining.
Several studies have evaluated the capabilities of generalist foundation models on medical tasks. Notably, ChatGPT-3.5 was able to pass the United States Medical Licensing Exam (USMLE) without specialized training~\cite{kung2023performance}. Similarly, GPT-4, using simple five-shot prompting, surpassed the USMLE passing score by over 20 points~\cite{nori2023capabilities}. The remarkable capabilities of these models have also been observed in medical exams conducted in non-English languages~\cite{kasai2023evaluating}.

To further improve the performance of generalist frontier models in medical challenge problems, the Medprompt strategy~\cite{nori2023can} was proposed. Medprompt integrates multiple advanced prompting techniques, including nearest-neighbor dynamic few-shot prompting and choice-shuffle ensembling. Notably, GPT-4's performance improved when it generated its own chain-of-thought reasoning for few-shot examples. Using Medprompt, GPT-4 achieves a MedQA score of 90.2\% and shows strong performance across various other medical benchmarks~\cite{nori2023can}.

While computationally expensive, fine-tuning LLMs for medical applications has also proven effective. Models like Med-PaLM~\cite{singhal2022large} and Med-PaLM 2~\cite{singhal2023towards}, which use the 540B-parameter Flan-PaLM with instruction prompt tuning, have achieved strong results in medical QA tasks. Med-PaLM was developed with a clinician-curated dataset, while Med-PaLM 2 used comprehensive instruction-based fine-tuning. Similarly, Med-Gemini, fine-tuned on multiple medical datasets and enhanced with web search during training and inference, achieved a 91.1\% score on MedQA \cite{saab2024capabilities}.

The recent introduction of OpenAI's o1 preview model~\cite{openai_o1_system_card_2024}, released in September 2024, marks another significant advancement. According to~\cite{openai_o1_system_card_2024}, the o1 preview model is trained using reinforcement learning to ``think'' before generating final responses. For challenging problems, o1 can dynamically increase computational resources during inference to achieve better results. 
While best practices for using the o1 model are still evolving, initial findings indicate that conventional approaches, such as explicit CoT prompting, may sometimes reduce its performance~\cite{openai2024reasoning}.

\subsection{Medprompt: Steering generalist models for specialized domains}

\begin{figure}[H]
    \centering
    \includegraphics[width=0.9\textwidth]{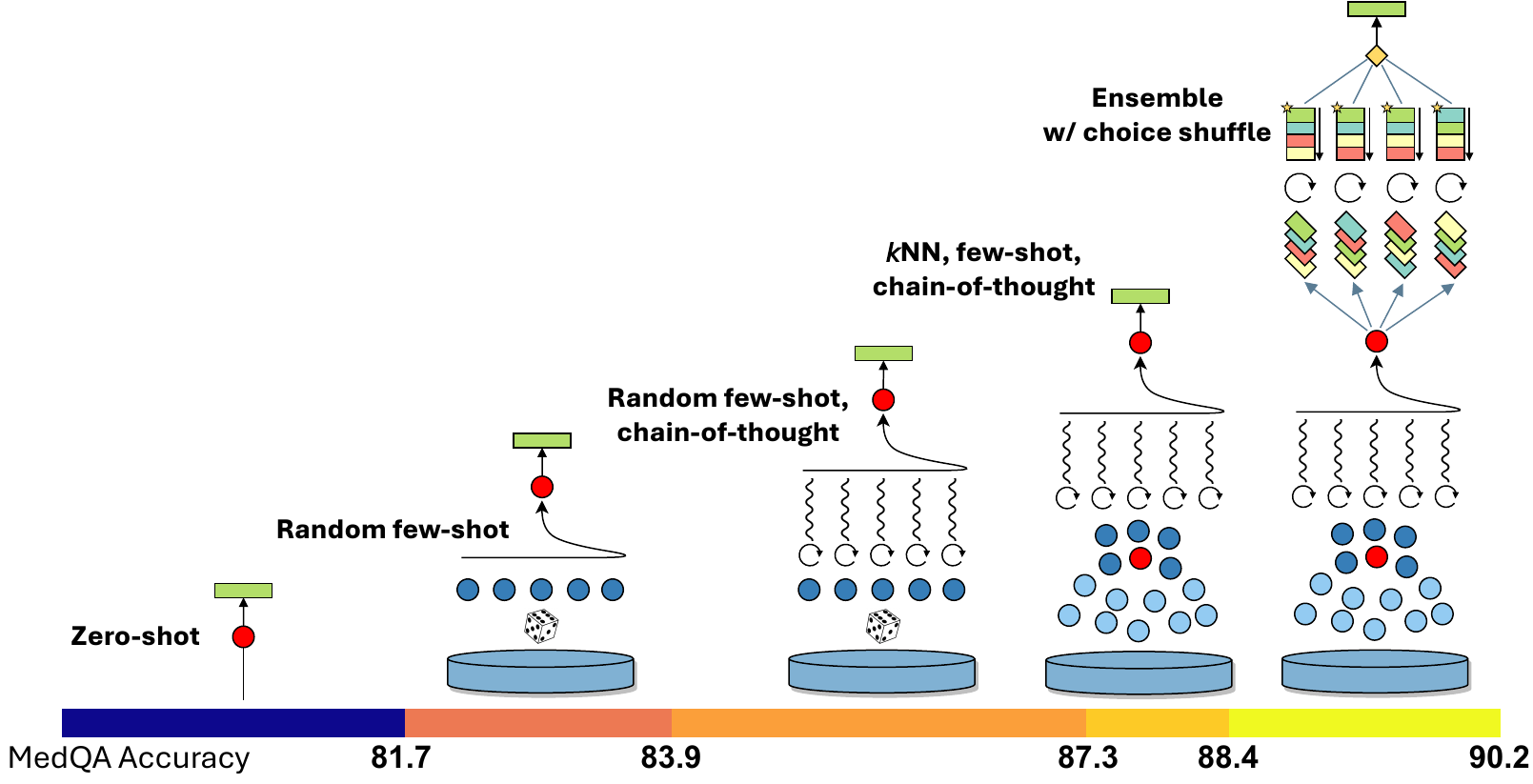}
    \caption{Visual illustration of Medprompt components and additive contributions to performance on MedQA. The prompting strategy combines $k$NN-based few-shot example selection, GPT-4--generated chain-of-thought prompting, and answer-choice shuffled ensembling.
    \newline Relative contributions of each component are shown at the bottom. Figure from \cite{nori2023can}}
    \label{fig:visual-medprompt}
\end{figure}

While frontier models like GPT-4 exhibit impressive general-purpose performance out-of-the-box, some specialized domains such as medicine require more adaptation for real-world utility. Medprompt offers a principled approach toward steering powerful general purpose models toward specialized domains at \emph{run-time} by combining several advanced prompting techniques to enhance performance in medical contexts. The framework consists of three main components: (1) dynamic, instance-specific in-context learning, (2) chain-of-thought reasoning, and (3) ensembling. As shown in Figure~\ref{fig:visual-medprompt}, each of these elements significantly improves performance~\cite{nori2023can}.

\subsubsection{Dynamic Few-Shot Prompting: Making the most of context} 

Language models exhibit flexible behavior depending on the content provided within their input context window, making prompting techniques essential for steering LLMs effectively. When solving a task, providing a clear task definition is important; otherwise, the model may become confused. Clear and precise instructions help narrow the text generation space, ensuring more focused and relevant responses.

An important technique for leveraging context is \emph{in-context learning} \cite{brown2020language}, where the model learns patterns and generalizes from a few examples provided at run-time. For instance, in medical question-answering (QA) tasks, a few relevant QA pairs can be presented to induce the desired answer style in response to a trailing question. This approach allows the model to adjust its token generation effectively based on examples---essentially adjusting model behavior at run-time by better specifying the context in which the model will operate. At the limit, dramatically increasing input tokens with thousands of few-shot examples can yield better performance at the cost of additional computational resources \cite{agarwal2024many}.

Another approach to building more useful context is to tap external resources, such as databases, search engines, and tools. For example, rather than relying on fixed or random sets of prompts, the user can retrieve relevant information from external resources, a technique often referred to as Retrieval-Augmented Generation (RAG). Medprompt performed RAG by searching databases for the most relevant examples to address each specific problem. Specifically, we employ text-embedding models in our studies with Medprompt: queries and examples are embedded into a shared semantic space, enabling the retrieval of closely aligned matches from the database.

The strategic use of context remains important, even for next-generation models like the o1 family. Clear, concise instructions, combined with dynamic access to tools, search capabilities, and databases, enable these models to build rich contextual understanding and improve responses.

\subsubsection{Chain of Thought: Using more tokens for reasoning} 

Chain-of-thought reasoning, which encourages the model to think step-by-step rather than generating direct answers, is also a central component of Medprompt \cite{wei2022chain}. For complex medical queries, breaking the problem into smaller, incremental steps improves accuracy. By either prompting (e.g., ``Let’s think step by step'') or using few-shot examples with explanations, we can guide LLMs to perform CoT reasoning. This technique can be seen as a run-time scaling strategy, as it requires more tokens, i.e., marshaling more compute to produce each answer.

One approach with Medprompt is to use few-shot examples with explanations to induce chain-of-thought reasoning. 
To prepare these few-shot examples, Medprompt employed GPT-4 to generate explanations for each question from the candidate pool. 
If GPT-4's explanations did not align with the ground truth answers, those candidates were removed from the pool. 
Interestingly, the GPT-4-generated explanations were more detailed than those prepared by human experts and proved to be more effective as few-shot examples.

The OpenAI o1 model is trained using reinforcement learning to reason internally before generating its final response \cite{o1-blog}. Previously, explicit prompting was necessary to induce this behavior. However, with o1, this strategy is internalized and can outperform external guidance. In fact, dedicated CoT prompting for o1 is unnecessary and officially not recommended by the model developers~\cite{openai_o1_guide_reasoning_2024}.

\subsubsection{Ensembling: Beyond a single LLM run} 

Another key direction is orchestrating multiple models, or multiple calls of the same model, to improve accuracy and reliability. One such example is \emph{ensembling}. Ensembling over portfolios of inferential chains has been employed in AI research for decades for projects focused on inference (such as studies in theorem proving) and machine learning. In machine learning, ensembling methods can reduce variance with little increase in bias, which also proves highly beneficial for language models \cite{caruana2004ensemble, wang2022self}. Medprompt applies ensembling by generating answers from multiple independent runs of the same question and using majority voting to determine the final output. This ensembling process aligns with the idea of different agents exploring diverse reasoning paths in parallel, followed by a consensus step to reach a conclusion.

To further enhance diversity in reasoning paths, Medprompt randomizes the order of multiple-choice options across runs. Language models, like humans, can exhibit slight biases toward the order of presented options \cite{blunch1984position, zheng2023judging, ko2020look}. Shuffling helps mitigate such positional biases. As a result, the aggregated output from these multiple runs is more robust, leading to improved accuracy and consistency.

Orchestrating multiple LLMs remains an important growth area for LLMs including o1. Emerging techniques such as multi-agent collaboration, discussions, consensus-building, and orchestration hold great potential to further enhance the capabilities of LLMs.

\section{Experimental Setup}

We evaluate the performance of o1-preview on a set of medical benchmarks and compare its accuracy with state-of-the-art models, including capabilities revealed via Medprompt. Our goal is to assess both the medical knowledge and reasoning ability of the models, particularly in scenarios involving patient cases. Our evaluation focuses on multiple-choice question (MCQ) formats, and we report accuracy as the primary evaluation metric for all datasets. We conduct experiments on medical benchmarks: MedQA~\cite{jin2021disease}, MedMCQA (Dev set)~\cite{pal2022medmcqa}, MMLU (Medical subset)~\cite{hendrycks2020measuring}, NCLEX (Nurse licensing exam)~\cite{nori2023can}, and the newly introduced JMLE-2024 benchmark~(Section \ref{sec:jmle-2024}). We also evaluate against two sets of official preparatory materials offered by the National Board of Medical Examiners (NBME) to help candidates prepare for the US Medical Licensing Exam (USMLE). For fair comparison with prior work, all experiments follow the benchmark setups described in our prior publication on Medprompt \cite{nori2023can}. 

\section{Results}

\subsection{Main Results}

Tables \ref{tab:main} and \ref{tab:usmle-exams-compact} present the performance of the o1-preview model and baseline models across the benchmarks. The results show that o1-preview achieves impressive performance in many tasks, surpassing the baseline GPT-4 models. Notably, the o1-preview numbers involve simple 0-shot prompting, in contrast to more sophisticated strategies like Medprompt. 

While o1-preview’s performance is impressive, it does not always outperform GPT-4 models. For example, MMLU Clinical Knowledge yield better results with Medprompt-enhanced GPT-4. OpenAI’s report corroborates this observation, noting that the effectiveness of o1-preview depends on the specific task setup~\cite{o1-blog}.
These findings suggest that prompting techniques are less necessary for the o1 model, but they remain valuable tools for traditional GPT-4 models in achieving high performance on specialized tasks.

\begin{table}[H]
\begin{center}
\begin{threeparttable}
\centering
\caption{Performance of different models on a suite of medical benchmarks.}

    \begin{tabular}{lccccc}
        \toprule %
        \multirow{2}{*}{Dataset}  & \multicolumn{1}{c}{GPT-4\tnote{*}} & \multicolumn{1}{c}{GPT-4-Turbo\tnote{**}} & \multicolumn{1}{c}{GPT-4o} & \multicolumn{1}{c}{GPT-4 Turbo\tnote{**}} & \multicolumn{1}{c}{o1-preview} \\ %
        & \multicolumn{1}{c}{(0 shot)} & \multicolumn{1}{c}{(5 shot)} & \multicolumn{1}{c}{(0 shot)} & \multicolumn{1}{c}{(Medprompt)} & \multicolumn{1}{c}{(0 shot)} \\
        \midrule %
        \textbf{MedQA (US 4-option)} & 78.9\% & 81.4\% & 84.4\% & 90.2\% & \textbf{96.0\%} \\
        \midrule
        \textbf{JMLE-2024} & - & 87.3\% & 92.7\% & 92.7\% & \textbf{98.2\%} \\
        \midrule %
        \textbf{MedMCQA Dev} & 69.5\% & 72.4\% & 76.4\% & 79.1\% & \textbf{83.9\%} \\
        \midrule %
        \textbf{MMLU} & & & & & \\ 
        Clinical Knowledge & 86.0\% & 86.4\% & 89.1\% & \textbf{95.8\%} & 93.6\% \\
        Medical Genetics & 91.0\% & 92.0\% & 96.0\% & 98.0\% & \textbf{99.0\%} \\
        Anatomy & 80.0\% & 80.0\% & 88.2\% & 89.6\% & \textbf{93.3\%} \\
        Professional Medicine & 93.0\% & 93.8\% & \textbf{97.4\%} & 95.2\% & 97.0\% \\
        College Biology & 95.1\% & 95.1\% & 95.1\% & 97.9\% & \textbf{98.6\%} \\ 
        College Medicine & 76.9\% & 76.9\% & 85.6\% & 89.0\% & \textbf{90.2\%} \\
        \bottomrule %
    \end{tabular}

\label{tab:main} %
\begin{tablenotes}
\item[*] Results originally reported in  \cite{nori2023capabilities} on the initial GPT-4 model release.
\item[**] Experimental results originally reported in \cite{nori2023can}, except JMLE-2024.
\end{tablenotes}
\end{threeparttable} \\
\end{center}
\end{table}

\begin{table}[H]
\begin{center}
\begin{threeparttable}
\centering
\caption{Comparative analysis of performance of different models on USMLE Sample Exam and USMLE Self Assessment.}

    \begin{tabular}{lcccc}
        \toprule %
        \multirow{2}{*}{Dataset}  & \multicolumn{1}{c}{GPT-4\tnote{*}} & \multicolumn{1}{c}{GPT-4o} & \multicolumn{1}{c}{o1-preview} \\ %
        & \multicolumn{1}{c}{(0 shot)} & \multicolumn{1}{c}{(0 shot)} & \multicolumn{1}{c}{(0 shot)} \\
        \midrule %
        \textbf{USMLE Sample Exam} & & & \\
        Step 1 & 80.7\% & 89.8\% & \textbf{91.6\%} \\
        Step 2 & 81.7\% & 89.1\% & \textbf{92.5\%} \\
        Step 3 & 89.8\% & 93.1\% & \textbf{96.4\%} \\
        \midrule %
        \textbf{USMLE Self Assessment} & & & \\
        Step 1 & 83.5\% & 87.5\% & \textbf{92.4\%} \\
        Step 2 & 84.8\% & 91.8\% & \textbf{93.4\%} \\
        Step 3 & \textbf{81.3\%} & 80.6\% & 80.7\% \\
        \bottomrule %
    \end{tabular}

\label{tab:usmle-exams-compact} %
\begin{tablenotes}
\item[*] Results reported originally n \cite{nori2023capabilities} on the initial GPT-4 model release.
\end{tablenotes}
\end{threeparttable} \\
\end{center}
\end{table}

\subsection{Performance on Newly Curated Multilingual Benchmark: JMLE-2024}
\label{sec:jmle-2024}

As with any study of public benchmarks on LLMs that consume massive amounts of training data, the question of test-set contamination and its influence on measured capabilities remains. For this study, we prepared a new benchmark, JMLE-2024, derived from the national medical licensing exam held in Japan in February 2024~\footnote{\url{https://www.mhlw.go.jp/seisakunitsuite/bunya/kenkou_iryou/iryou/topics/tp240424-01.html}}. This exam was administered after the knowledge cutoff date for the o1 model.

\begin{figure}[H]
    \centering
    \includegraphics[width=0.8\textwidth]{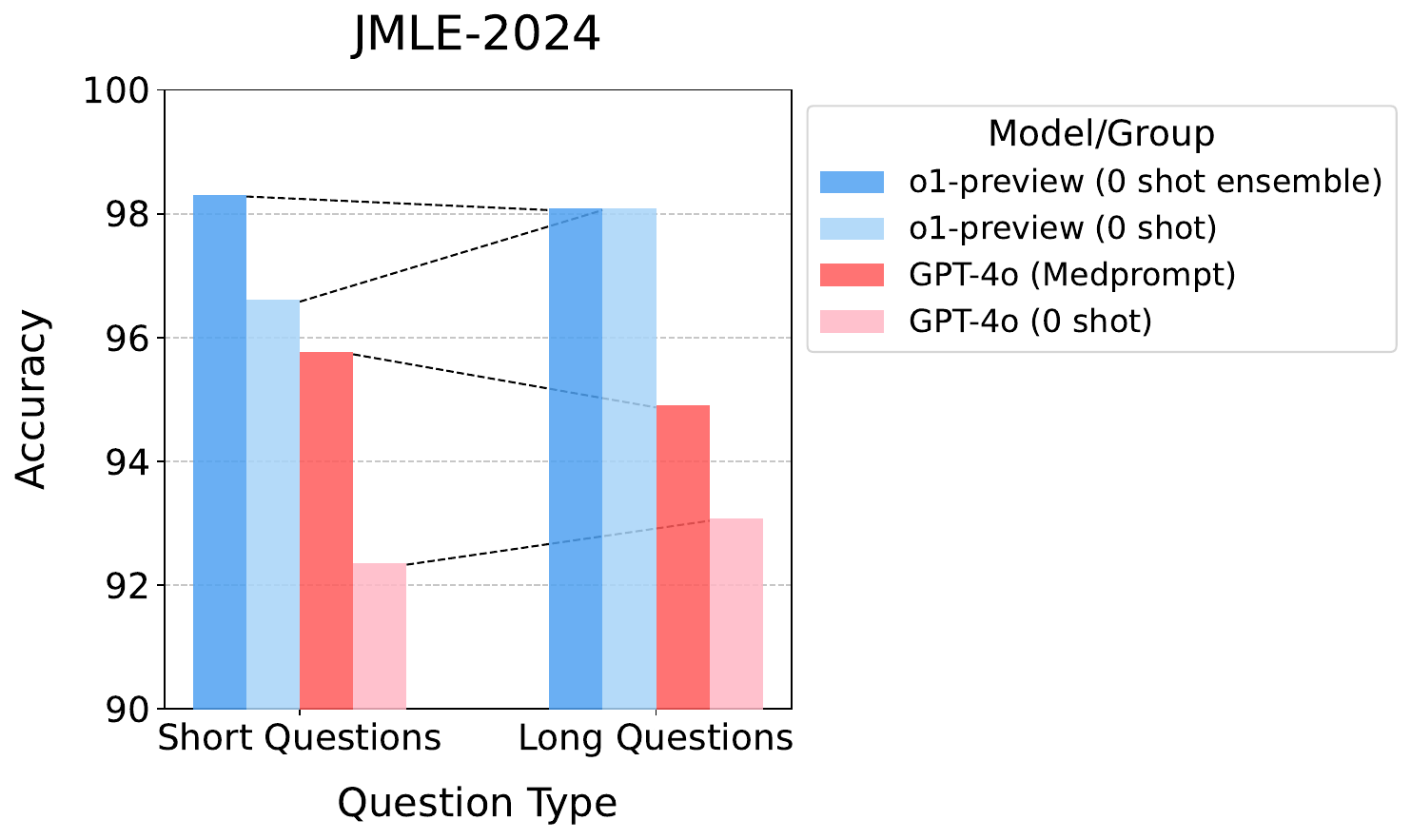}
    \caption{JMLE-2024: National medical licensing exam held in
Japan in February 2024}
    \label{fig:jmle}
\end{figure}

To construct the dataset, we filtered out questions containing images or those with multiple correct answers, retaining only five-choice questions with a single correct answer. After applying these criteria, the final dataset consisted of 275 questions. The questions spanned a broad range of medical topics, from fundamental clinical knowledge to complex case-based scenarios. The passing score for test takers on this exam is typically around 80\%. Although the Japanese medical licensing exam includes ``forbidden options,'' where selecting four or more such options results in automatic failure, we did not incorporate this criterion in our analysis.

We further divided the dataset into short and long question sets. Long questions typically include patient profiles and pose more intricate clinical problems. The results are presented in Figure~\ref{fig:jmle}. Consistent with other medical benchmarks, o1-preview demonstrates remarkable performance. Interestingly, o1-preview performed better on the long question set, further increasing the gap compared to the other baselines. This finding suggests that o1-preview can leverage its reasoning capabilities more effectively when answering more challenging problems. Notably, we did not translate the questions to English; instead, we directly used Japanese. Further performance gains were achieved using run-time steering techniques, such as Medprompt, along with ensemble approaches, which boosted the results for both GPT-4o and the o1-preview model.

While further evaluation is required, these results suggest: (1) o1-preview's strong performance is not merely based on memorization of well-established benchmarks, as the exam was made public after the model's knowledge cutoff date, and (2) o1-preview demonstrates remarkable proficiency in answering non-English medical questions.

\subsection{Impact of Prompting Techniques on o1-Preview Performance}

We now explore how different prompting strategies influence the performance of the o1-preview model. Our goal is to determine whether more advanced prompting methods provide additional benefits compared to simpler approaches with this new class of model.

Since the o1-preview model generates non-deterministic outputs by design (e.g., with a fixed sampling temperature of 1.0)~\cite{openai_o1_guide_reasoning_2024}, we conducted three independent runs for each evaluation to minimize the impact of stochastic variation. We report both the mean performance and standard deviation across these runs to ensure robustness of the results.

\begin{figure}[H]
    \centering
    \includegraphics[width=0.85\textwidth]{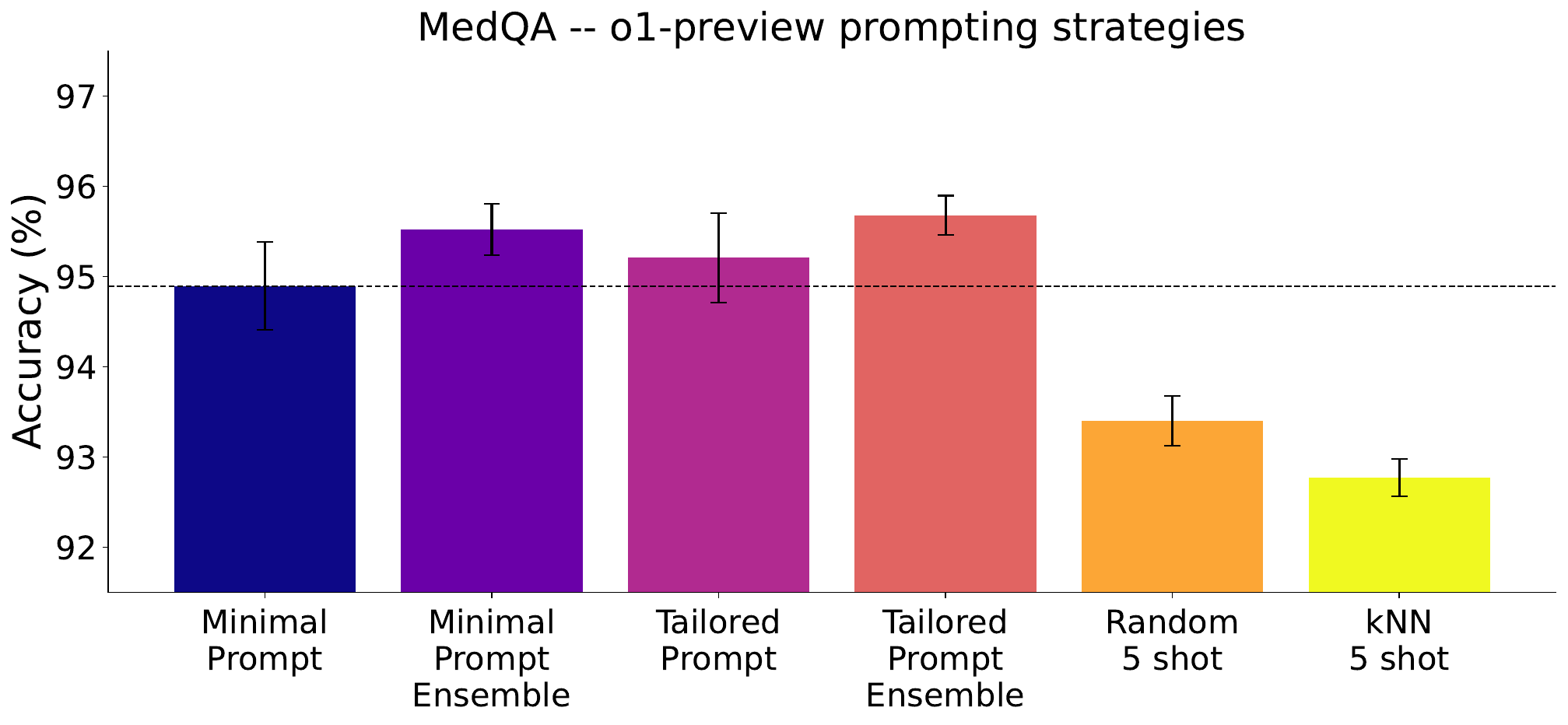}
    \caption{Comparison of prompting techniques on MedQA with the o1-preview model. Error bars indicate one standard deviation from three independent samples.}
    \label{fig:ablations}
\end{figure}

Figure~\ref{fig:ablations} presents the results on the MedQA dataset, including error bars to illustrate performance variability. Notably, five-shot prompting resulted in a significant \emph{decrease} in performance, suggesting that providing multiple similar examples in the prompt may have confused the model. This observation aligns with the OpenAI official guide, which notes that excessive retrieved context may impair performance~\cite{openai_o1_guide_reasoning_2024}.

\begin{figure}[H]
    \centering
    \includegraphics[width=0.95\textwidth]{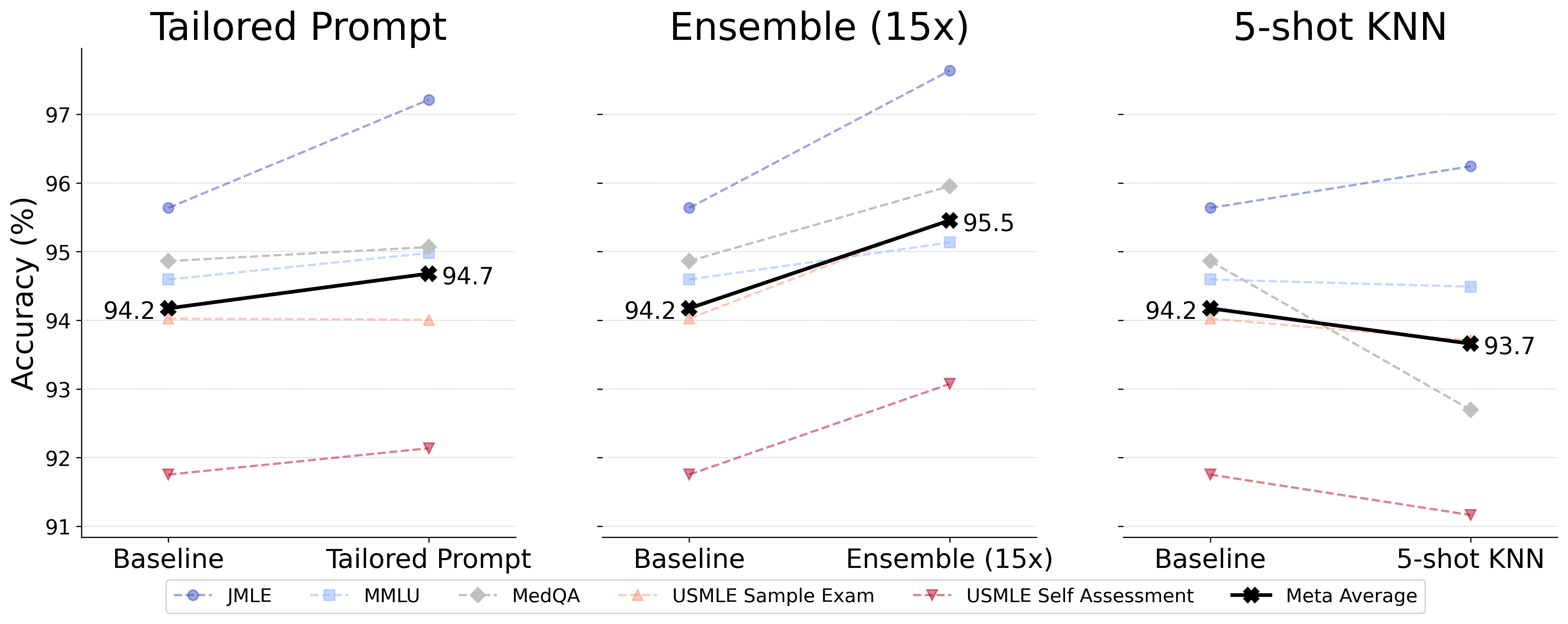}
    \caption{Tests of different prompting strategies across benchmark datasets. Ensembling outputs consistently yielded strong boosts in performance. Writing a custom-tailored prompt that describes the task in greater detail (Figures  \ref{fig:tailored-med},\ref{fig:tailored-MMLU}) had a marginal positive effect. Use of few-shot prompting was generally neutral or negative. Detailed results from all experiments are presented in Section \ref{sec:experiment-details}.}
    \label{fig:ablations-slope}
\end{figure}

In Figure ~\ref{fig:ablations-slope}, we individually test the efficacy of Medprompt's core components across the benchmark datasets. As seen with MedQA, tailored prompting has a small positive effect on performance. Few-shot prompting was inconsistent and, on average, hurt performance. Ensembling outputs across multiple runs, however, provided a consistent and meaningful boost across every benchmark task. We employed the same majority vote ensemble technique as \cite{nori2023can}, shuffling answer choices across each ensemble run to further increase diversity across the reasoning chains. As shown in the Medprompt and Self-Consistency prompting strategies \cite{nori2023can, wang2022self}, marginalizing over several diverse reasoning chains continues to be an effective variance reduction technique. 

\subsection{Role and Performance of Reasoning Tokens}

\begin{figure}[H]
    \centering
    \includegraphics[width=0.9\textwidth]{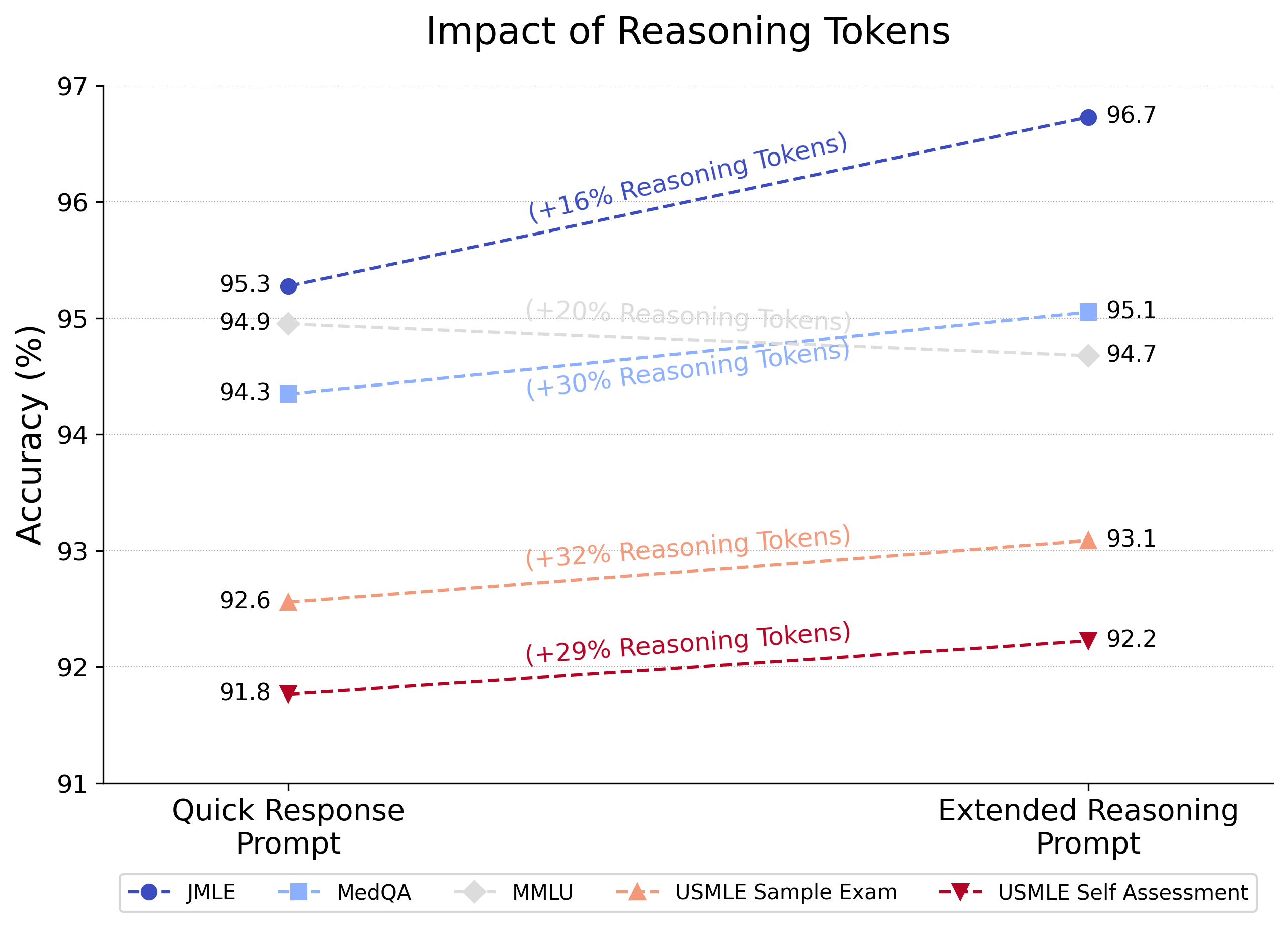}
    \caption{Effect of two prompting strategies which elicit variable length reasoning chains across benchmark datasets. Model accuracy tends to trend upwards when more reasoning tokens are used. Reasoning token count is returned from the OpenAI API. Exact prompt specifications are displayed in Figures \ref{prompt:speedy} and \ref{prompt:turtle}.}
    \label{fig:reasoning-scatterplot}
\end{figure}

The token usage of the o1-preview model is decomposed into three components: input tokens, representing the tokens used in the initial prompt; reasoning tokens, which are generated during intermediate reasoning steps; and output tokens, corresponding to the final response. We experimented with an alternative prompt format, instructing the model to provide only the final answer without accompanying explanations or rationales. This setup takes advantage of o1-preview's reasoning tokens. Even when instructed to output only the final answer, the o1-preview model can freely reason using its internal reasoning tokens. The results are presented in Figure~\ref{fig:reasoning-scatterplot}. 

OpenAI has suggested that leveraging higher numbers of reasoning tokens typically leads to higher performance \cite{o1-blog}. We corroborate this guidance with our experiments; we find that we can elicit significantly more reasoning tokens from the model purely via prompting, and that performance improves meaningfully when the model is explicitly told to spend longer amounts of time reasoning.

\subsection{Accuracy and Cost Frontiers}

We now explore tradeoffs between accuracy and cost using the MedQA benchmark. Evaluating the performance of LLMs involves balancing accuracy against cost. Techniques for run-time steering, such as Medprompt, can enhance output quality. However, they increase token consumption and, thus, raise inference costs. For instance, while the o1-preview model achieves excellent performance, it comes with higher operational expenses.

Figure~\ref{fig:cost} summarizes the total cost of processing 1,273 MedQA questions. Table~\ref{tab:cost-comparison} outlines the API pricing used in our analysis. Notably, o1-preview delivers superior accuracy at a lower cost compared to Medprompt with GPT-4 Turbo (November 2023). We note that GPT-4 Turbo (November 2023) powered the original Medprompt implementation~\cite{nori2023can} and was identified as a state-of-the-art solution at that time. Interestingly, o1-preview with minimal or few-shot prompting employs more tokens but performs worse than with tailored prompts, reinforcing the importance of prompt optimization. Additional ensembling further boosts the accuracy of o1-preview but this strategy also increases costs.

The latest model in the GPT-4 series, GPT-4o (August 2024), introduces a new balance between cost and accuracy. Compared to GPT-4 Turbo, GPT-4o consistently offers both improved accuracy and lower costs across all prompting strategies. As illustrated in Figure~\ref{fig:cost}, GPT-4o significantly shifts the cost-accuracy Pareto frontier, outperforming GPT-4 Turbo on both metrics. While o1-preview excels in terms of absolute accuracy, GPT-4o offers a well-balanced solution with strong performance at a lower cost. For example, increasing from $88\% \rightarrow 92\% \rightarrow 96$\% accuracy on the MedQA dataset each requires an order of magnitude increase in price (${\approx}\$5 \rightarrow  {\approx}\$50 \rightarrow {\approx}\$500$) with the GPT-4o Few-shot, GPT-4o Medprompt, and o1-preview (5x ensemble) strategies respectively. Depending on the specific use case and overall system setup, it is crucial to carefully consider cost-benefit tradeoffs and to optimize resources according to goals and preferences.

\begin{table}[H]
\begin{center}
\begin{threeparttable}
\centering
    \begin{tabular}{lccc}
        \toprule %
        \multirow{2}{*}{Model} & \multicolumn{3}{c}{Cost (\$) per 1M Tokens} \\ %
        \cmidrule(lr){2-4}
         & Input Tokens & Reasoning Tokens & Output Tokens \\        
        \midrule %
        o1-preview (Sep 2024) & 15 & 60 & 60 \\
        GPT-4o (Aug 2024) & 2.5 & - & 10 \\
        GPT-4 Turbo (Nov 2023) & 10 & - & 30 \\
        \bottomrule %
    \end{tabular}

\caption{API costs per 1M tokens used in the cost analysis. Cost is based on latest prices for generating tokens as of Oct 2024 for all models under consideration. Reasoning tokens are a new concept introduced in the o1 line of models. They are priced identically to output tokens.}
\label{tab:cost-comparison} %
\end{threeparttable} \\
\end{center}
\end{table}

\section{Directions with LLM Run-Time Strategies}

Scaling laws~\cite{kaplan2020scaling} for large language models (LLMs) have proven to be valuable for predicting how model capabilities grow with increasing data, compute, and model size. However, inference-time scaling---which focuses on the value of investing in additional computation at model run-time---is a promising new area of study. We see opportunities for innovating on how best to guide such run-time allocations for advancing efficiency, accuracy, and reasoning abilities. We review in this section promising directions in this emerging area of research and development, partly framed by our experimental results and prior research in AI on reasoning, ensembling, and control of the nature and extent of inference.

\subsection{Metareasoning principles and machinery} 

Prior research in AI on metareasoning---and in cognitive science on metacognition---can provide valuable insights into controlling and optimizing real-time deliberation over multiple generative threads in large language models (LLMs). High-level metareasoning methods can facilitate runtime decision making by dynamically allocating computing resources across different generative processes and their combinations. Specifically, these methods can incorporate considerations such as formulating and testing chains of thought, assessing confidence levels, estimating success likelihoods, and implementing control strategies to guide lower-level token-generation processes. 

Metareasoning methods drawn from the AI literature can be adapted for control and optimization to guiding run-time streams of token generation or can alternatively inspire more deeply woven, implicit representations that are learned from training data about problem solving. Metareasoning principles and mechanisms include methods to guide reasoning, execute portfolio strategies, and dynamically balance computational accuracy and cost. More broadly, context-sensitive control over cognitive resource allocation has been proposed as a unifying perspective on intelligence, bridging neuroscience, psychology, and AI research \cite{ScienceCompRational2015}. 

Relevant prior work includes optimizing the value of inference under varying constraints and uncertainties in allocated computing resources \cite{horvitz1987reasoning,horvitz1988reasoning} and use of estimates of solution likelihood or expected value of computation to determine when to continue, switch, or halt reasoning threads \cite{IJCAI89metareasoning,AIJ2001principles,UAI2001Satisfiability}. Other research has investigated models for ideal context-sensitive partition of time or other problem-solving resources to solution planning versus execution \cite{metapartition1990}.  Such opportunities for performing ideal ``metalevel partition'' can be valuable for building ideal run-time inference systems that maximize the efficiency or likelihood of success of base-level generation processes in LLMs. 

\subsection{Optimizing Input for LLMs} 

\subsubsection{In-Context Learning}

The in-context learning paradigm, introduced by \cite{brown2020language}, showed that extending model prompts with several demonstration examples---commonly referred to as few-shot prompting---significantly improves task accuracy across a wide range of challenges. Advancements in hardware and algorithms are enabling long-context modeling, but fully harnessing the potential of extended contexts remains a key challenge. \cite{agarwal2024many} explored the paradigm at scale, showing that language models continue to benefit from thousands of demonstration examples. OpenAI’s guidelines suggest that few-shot prompting or incorporating additional retrieved context may not always improve performance and, in some cases, might degrade it~\cite{openai_o1_guide_reasoning_2024}. Determining how to efficiently provide relevant examples and additional context to optimize performance---especially in models such as o1-preview---remains a promising area of research.

\subsubsection{Integrating External Resources at Runtime}

The progression of LLMs, such as the o1-preview model, has markedly reduced the dependence on intricate prompt engineering by baking reasoning behavior into the model's standard mode of operation. However, an essential avenue for further enhancing these models lies in their ability to actively acquire information at run-time from external sources such as the web and knowledge bases (KBs)~\cite{lewis2020retrieval}. 
The integration of active information acquisition has potential in several aspects.

By leveraging external databases, models can effectively bypass the limitations of their fixed training data, accessing a virtually unlimited pool of information. Additionally, models can also remain up-to-date without the need for frequent retraining, as they can incorporate the latest information from external sources, which is particularly beneficial in rapidly evolving fields like medicine or technology. There is opportunity for run-time inference to compute estimations of the expected value of information for different types of information.

Allowing models to leverage tools, typically in the form of software libraries and APIs, is another form of effective inference time compute scaling ~\cite{schick2023toolformer, bubeck2023sparks, patil2023gorilla}.

\subsection{Guiding LLM Inference and Sampling} 

New tools are emerging to enable finer-grained control and scaling of language model inference. Innovations in sampling methods, like entropy based sampling techniques~\cite{entropix}, show promise in better leveraging the inherent calibrated uncertainty a model assigns to each token as it generates text ~\cite{openai2023gpt4, nori2023capabilities}. Software tooling like Guidance ~\cite{guidance2024} and Outlines~\cite{willard2023efficient} allow for model developers to dynamically manipulate per-token probabilities, leading to higher quality outputs when language model inference is tightly coupled with external systems~\cite{hartvigsen2022toxigen}. While these tools currently operate without tight coupling to a model, incorporating this style of token steering mechanisms directly into model training may unlock further capabilities.

\subsection{Reasoning}

Chain-of-Thought prompting, introduced by \cite{kojima2022large, wei2022chain}, promotes step-by-step, auto-regressive token generation. The method has been demonstrated to significantly improve performance on tasks that appear to benefit from more complex steps of reasoning. The CoT technique encourages LLMs to break down their thought processes, improving the accuracy of intermediate steps, which in turn boosts final outcomes. Remarkably, \cite{pfau2024let} found that this improvement arises even when models produce meaningless intermediate tokens, suggesting that simply expending computational resources can enhance the reasoning performance of transformer-based models.

Recent research demonstrates that advanced prompting compositions further refine reasoning abilities. Methods such as ReAct~\cite{yao2022react}, skeleton-based prompting~\cite{ning2023skeleton}, and tree-based reasoning~\cite{yao2024tree} allow for more structured problem-solving, improving outcomes in multi-step tasks.

Training LLMs to improve real-time reasoning capabilities remains a highly active research area. The Self-Taught Reasoner (STaR) framework, introduced by ~\cite{zelikman2022star, zelikman2024quiet}, exemplifies this trend. STaR iteratively refines the model’s thought process by generating multiple reasoning paths (rationales), filtering them for correctness, and fine-tuning the model using successful outcomes. This self-improvement cycle allows models to enhance their reasoning abilities progressively.

Another notable framework, Let’s Verify Step-by-Step \cite{pfau2024let}, introduces process supervision to improve mathematical reasoning. By focusing on step-by-step validation rather than just output correctness, it improves the reliability of reward models, fostering better outcomes on tasks with high logical complexity.

A growing trend emphasizes scaling test-time computation over merely increasing model size or pre-training compute. Research by~\cite{snell2024scaling, wu2024empirical, wu2024thinking} highlights that leveraging iterative reasoning strategies at inference time can significantly enhance performance without the need for larger models.

OpenAI’s latest model, o1, exemplifies this trend by employing reinforcement learning (RL) to improve reasoning capabilities. Trained to think before responding, o1 delivers exceptional performance on reasoning-intensive tasks. 

\subsection{Leveraging Multiple Runs and Models}
\subsubsection{Ensembling}

Ensembling is a powerful technique in machine learning that enhances model performance by combining multiple models or their outputs~\cite{caruana2004ensemble}. It has also demonstrated promising results for large language models (LLMs)~\cite{wang2022self,nori2023can}.

While simple majority voting is a popular approach for aggregating outputs, more sophisticated methods are emerging. One such method is Ensemble Refinement, introduced with Med-PaLM 2~\cite{singhal2023towards}. This technique employs multi-stage aggregation by generating multiple reasoning paths through stochastic sampling. Rather than relying solely on majority voting, the model iteratively re-conditions on intermediate outputs, refining its reasoning at each stage to achieve higher precision.

Another approach, LLM-Blender~\cite{jiang2023llm}, incorporates two key components: a ranker and a fuser. The ranker performs pairwise evaluations of outputs from multiple models, identifying the most promising candidates. The fuser then merges these candidates into a coherent response, leveraging the strengths of each for an optimal outcome.

However, a major challenge with ensembling is the computational cost. One direction addressing this issue is dynamic self-consistency~\cite{wan2024dynamic,wang2024make}, which optimizes the sampling process by stopping early when sufficient consistency is detected among reasoning paths.

Future research is needed for developing adaptive ensembling strategies and further optimizing sampling methods to maximize performance while maintaining computational efficiency.

\subsubsection{Model Federation and Multi-Agent Architectures}

Recent research has investigated the use of agent frameworks and multi-agent orchestration, wherein models are equipped with access to various tools and collaborate to achieve state-of-the-art performance on complex, real-world tasks. These frameworks enable models to dynamically select and integrate the tools required to solve a given problem, thereby distributing reasoning and computation across specialized agents. \cite{wu2023autogen} presents an example of this orchestration, wherein models collaborate to leverage their respective strengths to accomplish intricate tasks. These multi-agent frameworks have yielded breakthrough performance on real-world tasks, beyond what could be elicited from a single model call alone \cite{jimenez2023swe}.

\section{Limitations}

We evaluated o1-preview on a set of medical benchmarks. We found that o1-preview significantly outperforms GPT-4 guided by Medprompt, a set of advanced dynamic prompting strategies that had previously achieved state-of-the-art performance on the medical benchmarks at focus. We note that this study is preliminary and limited in scope, and more extensive evaluations are necessary to fully understand o1-preview's capabilities. While the initial results are impressive, several challenges and opportunities for growth remain that warrant further discussions.

\paragraph{Benchmark Saturation.} One significant issue is the rapid saturation of existing benchmarks, limiting their utility in evaluating state-of-the-art models. In the MedGemini paper~\cite{saab2024capabilities}, Google researchers analyzed the MedQA dataset and found that a small but notable portion of the questions had labeling errors, missing information, or ambiguities. The effective performance of o1-preview on MedQA might have reached this dataset's noise ceiling. We anticipate similar issues in other benchmarks. We need to develop new benchmarks that are more challenging and relevant to real-world medical challenges.

\paragraph{o1-preview API Limitations.} Another area for consideration involves the limitations of the o1-preview  API~\cite{openai_o1_guide_reasoning_2024}. While o1 exhibits strong reasoning abilities by ``thinking'' before outputting responses, its internal planning and chain-of-thought processes are hidden, so we cannot report on the nature of these tokens. In addition, the o1 API currently exposes fewer sampling parameters (such as temperature, Top P) or custom system prompts, so we cannot yet explore these variables. 

\paragraph{Best Practices with o1.} Optimizing interactions with o1 remains an ongoing challenge. Our experimentation with various prompting techniques---some of which had previously yielded significant improvements with GPT-4--- revealed mixed results. Certain techniques provided no noticeable gains or, in some cases, even degraded performance when applied to o1-preview. Given that o1-preview is still a relatively new model, this opens up a valuable opportunity for further exploration. Identifying optimal interaction strategies represents an exciting area for research, as refining these best practices will be instrumental in helping users unlock the model's full potential.

\paragraph{Ethical Use.} Finally, the rapid development of models like o1-preview raises important questions about safety, ethical considerations, and responsible use. It is crucial to not only develop advanced capabilities but also ensure they are deployed in a manner that is safe, transparent, and beneficial for end users. Addressing these aspects will be essential as we move forward with refining and applying the o1 family of models in the medical domain.

\section{Conclusions}

We expect the landscape of run-time strategies for LLMs to evolve rapidly. We systematically evaluated OpenAI's o1-preview model on a set of medical benchmarks. Remarkably, even without advanced prompting techniques, o1-preview outperformed GPT-4 with Medprompt. Medprompt was initially designed to guide generalist models like GPT-4 in specialized domains using methods such as self-generated CoT and few-shot prompting. However, our results indicate that the o1-preview model reduces reliance on some of these techniques. In some cases, few-shot prompting even hindered o1-preview's performance, suggesting that in-context learning is not an effective way to steer these models.

While ensembling remains a viable approach, it is resource-intensive and requires careful optimization to balance costs and performance gains. Our cost and accuracy analysis of various run-time strategies revealed a new Pareto frontier, illustrating the trade-offs between cost-efficiency and accuracy. Although GPT-4o with steering strategies like Medprompt retains value in specific contexts depending on task requirements, optimizing run-time strategies for advanced models like o1-preview will require nuanced balance between control, performance, and resource allocation.

Moreover, our findings highlight that the o1-preview model has achieved near-saturation on many existing medical benchmarks. This study primarily utilized multiple-choice questions to assess knowledge and reasoning, which, despite their utility, have inherent limitations. Consequently, there is an urgent need for new, more challenging benchmarks to accurately evaluate these advanced models. Further evaluation of the o1-preview model in real-world applications and across diverse aspects of medical understanding will be essential.

\section*{Acknowledgments}

We thank Rich Caruana, Hoifung Poon, and Paul Vozila for insightful discussions and feedback.

\newpage

\bibliographystyle{alpha}
\bibliography{mainbib}

\section{Appendix}
\subsection{Full Experimental Results}
\label{sec:experiment-details}

\begin{figure}[H]
    \centering
    \includegraphics[width=0.9\linewidth]{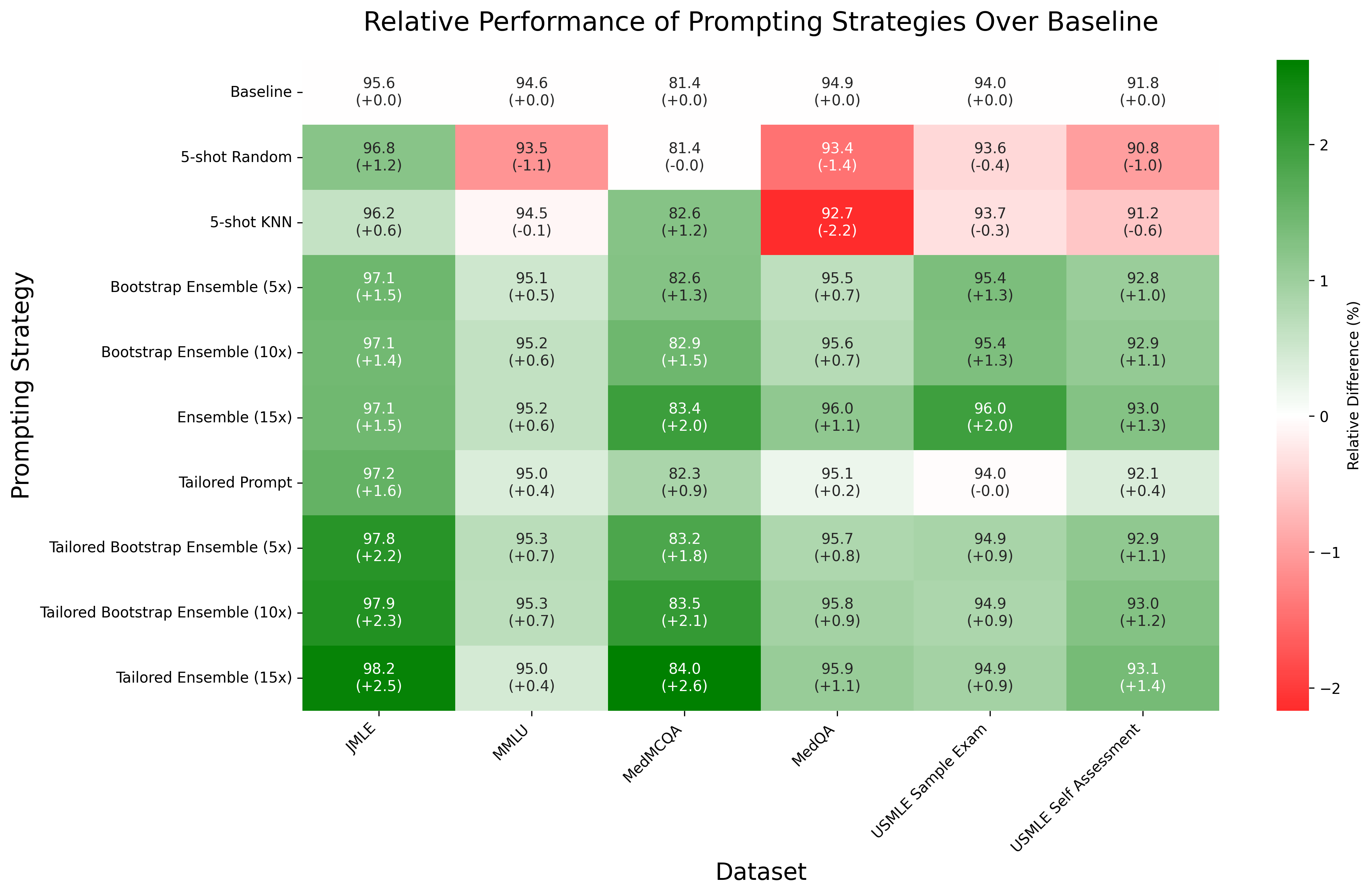}
    \caption{Heatmap showing absolute accuracy and relative performance over baseline zero-shot prompt (in parenthesis) across all benchmark datasets. As shown in Figure \ref{fig:ablations-slope},  ensembling strategies consistently helped performance, while few-shot produced mixed results.}
    \label{fig:heatmap}
\end{figure}

\subsection{Prompt Templates}

\begin{figure}[H]
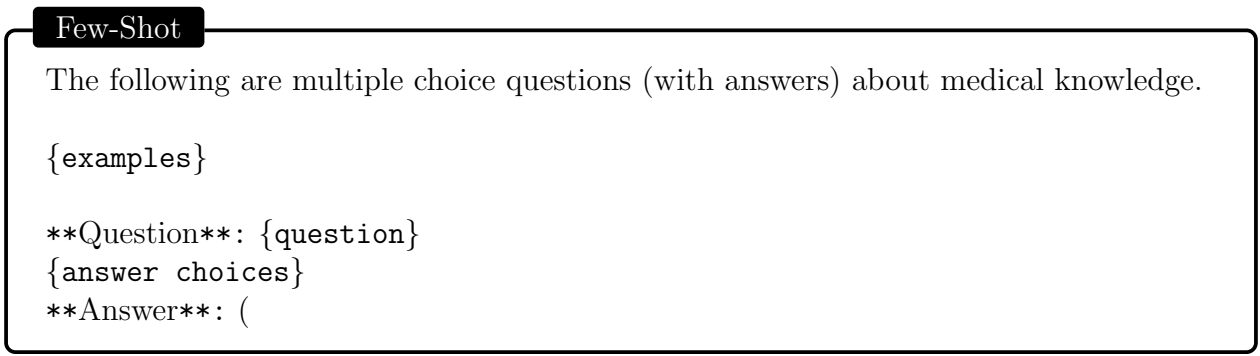

\begin{AIbox}{Few-Shot}
The following are multiple choice questions (with answers) about medical knowledge. \\
\\
\texttt{\{examples\}} \\
\\
\verb|**|Question\verb|**:| \texttt{\{question\}} \\
\texttt{\{answer choices\}} \\
\verb|**|Answer\verb|**:| (
\end{AIbox}
\caption{Template used for zero-shot and five-shot prompts.}
\end{figure}

\begin{figure}[H]
\begin{AIbox}{Quick Response}
Please answer the following question as quickly as possible. We have narrowed down the possibilities to four different answers. I am in an emergency, and speed is of utmost importance. It is more important to answer quickly than it is to analyze too carefully. Return just the answer as quickly as possible. \\

\texttt{------} \\
\# QUESTION \\

\texttt{\{question\}} \\

\# ANSWER CHOICES \\

\texttt{\{answer choices\}} \\
\texttt{------}
\\
Please remember to answer quickly and succinctly. Time is of the essence!
\end{AIbox}
\caption{Prompt that elicits the model to respond with less reasoning and completion tokens.}
\label{prompt:speedy}
\end{figure}

\begin{figure}[H]
\begin{AIbox}{Extended Reasoning}
Please answer the following multiple choice question. Take your time and think as carefully and methodically about the problem as you need to. I am not in a rush for the best answer; I would like you to spend as much time as you need studying the problem. When you're done, return only the answer. \\

\texttt{------} \\
\# QUESTION \\

\texttt{\{question\}} \\

\# ANSWER CHOICES \\

\texttt{\{answer choices\}} \\
\texttt{------}
\\
Remember, think carefully and deliberately about the problem. Take as much time as you need. I will be very sad if you answer quickly and get it wrong.
\end{AIbox}
\caption{Prompt that elicits the model to respond with more reasoning and completion tokens.}
\label{prompt:turtle}
\end{figure}

\begin{figure}[H]
\begin{AIbox}{Tailored - Medical Questions}
You are tasked with solving complex medical questions that assess both the knowledge and clinical reasoning required for a medical licensing exam. These questions cover critical topics such as anatomy, physiology, pathology, pharmacology, and patient management. Read the following question carefully and select the most accurate answer from the provided options. \\

\verb|**|Question\verb|**:| \\
\texttt{\{question\}} \\

\verb|**|Options\verb|**:| \\
\texttt{\{answer choices\}} \\

\verb|**|Instructions\verb|**:| \\
- Think deeply and thoroughly, then choose the best possible answer from the given options (only one choice). \\
- Your final response must contain only the letter corresponding to the correct answer (e.g., "A"). Do not include explanations or additional text in your output. \\

\verb|**|Answer\verb|**:|
\end{AIbox}
\caption{Prompt for medical questions requiring knowledge and clinical reasoning.}
\label{fig:tailored-med}
\end{figure}

\begin{figure}[H]
\begin{AIbox}{Tailored - MMLU}
You are tasked with answering academic questions from various subjects, including medicine, clinical knowledge, biology, anatomy, and more. Carefully read the following question and select the most accurate answer from the provided options. \\

\verb|**|Question\verb|**:| \\
\texttt{\{question\}} \\

\verb|**|Options\verb|**:| \\
\texttt{\{answer choices\}} \\

\verb|**|Instructions\verb|**:| \\
- Think deeply and thoroughly, then choose the best possible answer from the given options (only one choice). \\
- Your final response must contain only the letter corresponding to the correct answer (e.g., "A"). Do not include explanations or additional text in your output. \\

\verb|**|Answer\verb|**:|
\end{AIbox}
\caption{Prompt for MMLU dataset questions requiring accurate academic responses across multiple subjects.}
\label{fig:tailored-MMLU}
\end{figure}

\end{document}